\documentclass[runningheads]{llncs}

\usepackage{eccv}

\usepackage{eccvabbrv}

\usepackage{graphicx}
\usepackage{booktabs}

\usepackage[accsupp]{axessibility}  %

\usepackage{hyperref}

\usepackage{orcidlink}

\usepackage{bm}
\usepackage{multirow}
\usepackage[normalem]{ulem}
\useunder{\uline}{\ul}{}
\usepackage[noEnd=false, beginLComment=//~,endLComment=,beginComment=//~]{algpseudocodex}
\usepackage{algorithm}
\usepackage{wrapfig}
\usepackage{colortbl}
\usepackage{array}
\usepackage{colortbl}

\newcolumntype{C}[1]{>{\centering\arraybackslash}m{#1}}
\newcolumntype{L}[1]{>{\arraybackslash}m{#1}}

\begin{document}

\title{Fast Sprite Decomposition from Animated Graphics}

\author{Tomoyuki Suzuki\orcidlink{0000-0003-4273-2325} \and
Kotaro Kikuchi\orcidlink{0000-0003-1747-5945} \and
Kota Yamaguchi\orcidlink{0000-0002-3597-2913}}

\authorrunning{T.~Suzuki et al.}

\institute{CyberAgent
\\\email{\{suzuki\_tomoyuki,kikuchi\_kotaro\_xa,yamaguchi\_kota\}@cyberagent.co.jp}
}

\maketitle
\begin{abstract}
This paper presents an approach to decomposing animated graphics into sprites, a set of basic elements or layers.
Our approach builds on the optimization of sprite parameters to fit the raster video.
For efficiency, we assume static textures for sprites to reduce the search space while preventing artifacts using a texture prior model.
To further speed up the optimization, we introduce the initialization of the sprite parameters utilizing a pre-trained video object segmentation model and user input of single frame annotations.
For our study, we construct the Crello Animation dataset from an online design service and define quantitative metrics to measure the quality of the extracted sprites.
Experiments show that our method significantly outperforms baselines for similar decomposition tasks in terms of the quality/efficiency tradeoff.
\footnote{Project page: \url{https://cyberagentailab.github.io/sprite-decompose}}

\keywords{sprite decomposition \and animated graphics \and optimization}
\end{abstract}

\begin{figure}[t]
  \centering
  \includegraphics[width=\linewidth]{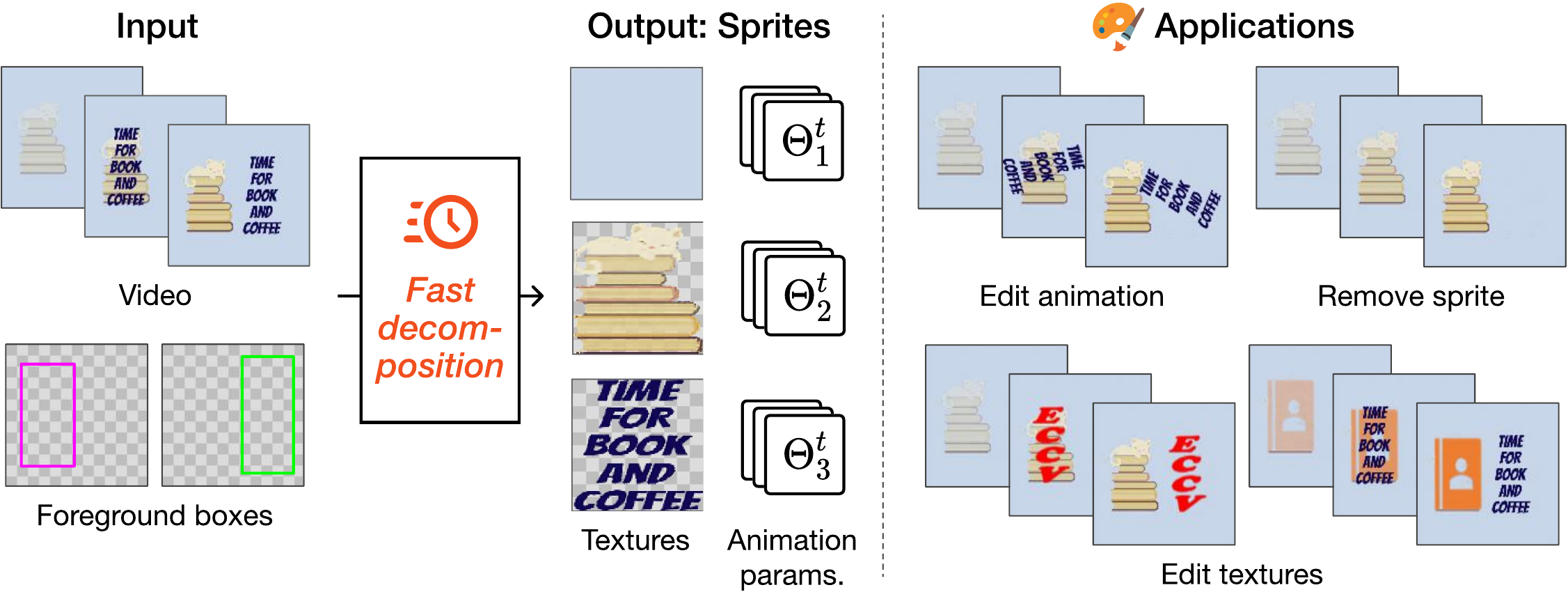}
  \caption{Sprite decomposition from animated graphics. Given a raster video and auxiliary bounding box annotations, our method decomposes sprites that consist of static textures and animation parameters. The decomposed parameters are easily applicable to various video-editing applications.}
  \label{fig:overview}
\end{figure}

\section{Introduction}
\label{sec:introduction}

Designers build animated graphics such as social media posts or advertisements using \emph{sprites}, a basic set of animated objects or layers in video editing.
Sprite allows intuitive manipulation of objects in videos thanks to its compact and interpretable data representation.
However, once a video is composited into a raster video through a rendering engine, it is almost impossible to manipulate objects in a video instantly.
Still, it is common to observe a situation where one wishes to edit certain parts of the raster animated graphic video, for example, when a user attempts to create an original from a reference raster animated graphic video.
This is where the decomposition of a raster video into the sprites comes into play.

In this work, we tackle the decomposition of raster animated graphics into sprites for video editing applications.
While there has been literature on video decomposition~\cite{lna,ds,zhang2022sprite}, we argue that the decomposition of animated graphics poses unique challenges.
Compared to the decomposition of natural scenes, animated graphics include more objects in a single video and involve many types of elements, such as background images, illustrations, typographic elements, or embellishments.
Each element has different dynamics, which are typically defined as animation effects (\eg, zoom-in or fade-out) in a video authoring tool.
While the dynamics or texture of the objects are usually simpler than natural scenes, any artifact resulting from the failure of decomposition is perceptually unacceptable for video editing applications.
For ease of manipulation, the animation parameters should be a compact parametric representation (in our case, affine transformation)
.
Also, considering that the users are designers, it is important to make a decomposition approach fast enough to implement in an interactive editing tool; \eg, it is not acceptable to spend hours processing a raster video in a workflow. %

Our sprite decomposition approach is designed to address the aforementioned challenges.
\cref{fig:overview} illustrates the overview of our decomposition task.
We formulate the sprite decomposition as an optimization problem to fit the sprite parameters to the given raster video.
Considering the typical scenario in animated graphics, we introduce a static sprite assumption that all the textures are static and only animation parameters change over time, significantly reducing the parameter space in the optimization process.
Under this assumption, we incorporate an image-prior model to prevent undesirable pixel artifacts.
For an efficient optimization process, we employ a gradient-based optimizer with an effective initialization procedure that builds on a video object segmentation model from minimal user annotation of object bounding boxes in a single video frame.
Combining those simple yet effective approaches, we achieve much faster convergence in the decomposition, which we show in the experiments.

To evaluate the quality and speed of animated graphics decomposition, we build \emph{Crello Animation} dataset that consists of high-quality templates of animated graphic designs, which we collect from an online design service.
Also, we define benchmark metrics to evaluate the quality of decomposition tailored for animated graphics.
In the experiment, we show our approach considerably outperforms similar decomposition baselines regarding the trade-off between quality and efficiency.
Finally, we present application examples of video editing using decomposed sprites by our approach.
We summarize our contributions as follows:
\begin{enumerate}
  \item We propose a simple and efficient optimization-based method for decomposing sprites from animated graphics.
  \item We construct the Crello Animation dataset and benchmark metrics to evaluate the quality of sprite decomposition.
  \item We empirically show that our method constitutes a strong baseline in terms of the quality/efficiency trade-off and achieves significantly faster convergence to reach the same decomposition quality.
\end{enumerate}

\section{Related work}
\label{sec:related_work}

\subsection{Image vectorization and decomposition}

Vectorizing or decomposing images is the inverse problem of rasterizing or rendering,
\ie, the task of converting an input image into a parametric representation that can be rasterized or rendered as visually identical to the input.
Motivations behind this task include editing raster content and obtaining scalable vector representations.
Several studies have been made in the computer vision and computer graphics communities for representations such as image layers~\cite{aksoy2017unmixing,sbai2022icip}, vector graphics~\cite{Li2020Differentiable,xu2022live}, and text attributes~\cite{Shimoda_2021_ICCV}.
A common approach is to use gradient-based optimization to search for parameters that minimize reconstruction error.
While we share the motivation and general approach with the above studies,
our work is differentiated by a new data representation, which we refer to as animated graphics, and a method designed specifically for this purpose.

\subsection{Video decomposition}

There have been many attempts to decompose a raster video into a sequence of layers. 
Omnimatte~\cite{lu2021omnimatte}, Layered Neural Rendering~\cite{geng2023learning}, DyStaB~\cite{yang2021dystab}, Double-DIP~\cite{gandelsman2019double} and amodal video object segmentation~\cite{xie2022segmenting,lamdouar2021segmenting} aim to decompose a video into layers (pixel arrays or masks), but they do not aim to parameterize them.

Wang and Adelson~\cite{wang1994representing} proposed a method to represent a layer as a pair of appearance and parameterized animation in addition to decomposing a video into layers, and following attempts~\cite{brostow1999motion,jojic2001learning,pawan2008learning,agarwal2003what,rav2008unwrap} have been made.
While the basic formulation has not changed since these early works, recent studies adopt machine learning approaches to the decomposition pipeline for better quality.
Layered Neural Atlases (LNA)~\cite{lna} represent primary objects in a video as 2D atlases and their dynamics as moving reference points in the 2D atlases.
The coordinate-based multilayer perceptrons (MLPs)~\cite{chen2021learning,sitzmann2020implicit,tancik2020fourier} represent the 2D atlases and the mapping of reference points.
Lee~\etal~\cite{lee2023shape} extends LNA to edit the appearance of the atlas based on text prompts and compensate for changes in deformation through estimated semantic correspondences.
Deformable Sprites (DS)~\cite{ds} is similar to LNA but differs in that the 2D atlas (or texture) is simplified to a pixel grid, and deformations are parameterized as B-splines.
Sprite-from-Sprite~\cite{zhang2022sprite} decomposes a cartoon into sprites, where each sprite is represented by a homography warping and spatiotemporal pixel grid including all other information.
In this representation, it is difficult to propagate the appearance manipulation temporally, unlike LNA, DS and ours.

\begin{figure}[t]
  \centering
  \includegraphics[keepaspectratio,width=0.98\linewidth]{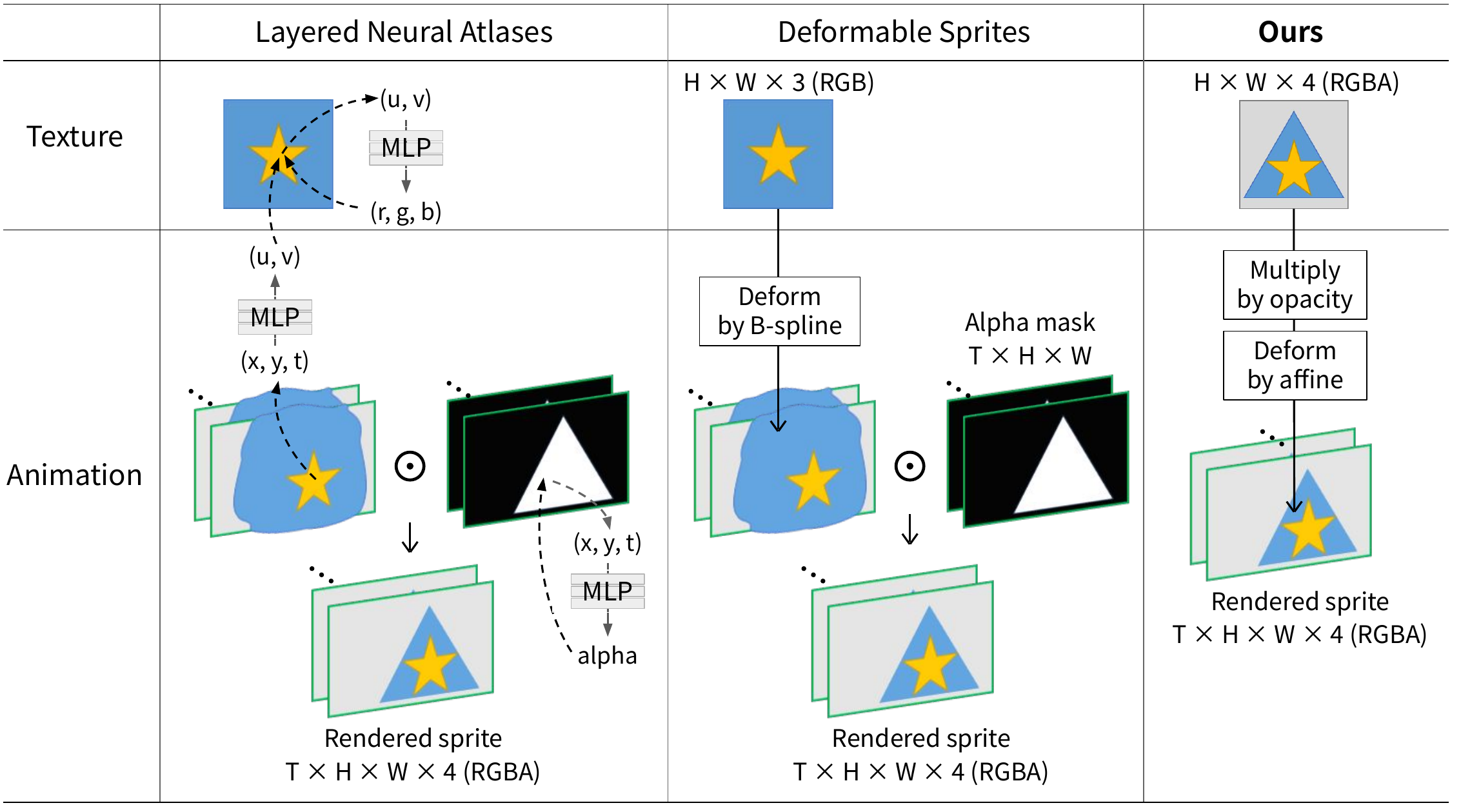}
  \caption{Comparison of sprite representations of Layered Neural Atlases~\cite{lna}, Deformable Sprites~\cite{ds}, and ours. Our approach limits parameter space to static texture and affine transformation, which enables faster convergence while keeping the necessary representation for animated graphics.}
  \label{fig:comparison_lna_ds}
\end{figure}

We compare our method with LNA~\cite{lna} and DS~\cite{ds},
which have static textures in their sprite representation like ours.
The detailed differences in sprite representation are summarized in \cref{fig:comparison_lna_ds}.
Compared to these methods, our method has a minimal yet sufficient parametrization to cover the typical cases of animated graphics and is particularly beneficial when editing textures.
As discussed in~\cite{lee2023shape}, LNA requires correction of the mapping between the original and edited textures.
The B-spline transform used in DS is flexible for general video decomposition but can result in unwanted deformation for our animated graphics.
Both methods use only pixel-level alpha masks, which may not capture the temporal changes in sprite-level opacity often seen in fade-in and fade-out animations.
Our simplified representation also leads to faster convergence, is further accelerated by our dedicated initialization and gives a prior for better decomposition quality.

Another line of approach is to decompose and parameterize a video using auto-encoder-based disentangled representation learning~\cite{smirnov2021marionette,monnier2021dtisprites}.
However, these methods assume that a certain amount of training data is available in the target domain, while we only require the target video for optimization.

\section{Sprite decomposition}
\label{sec:approach}

\subsection{Data definition}
\label{sec:data_definition}
In this work, we define an animated graphic $X$ as a sequence of $K$ sprites, each consisting of a static texture image $\bm{x}_k$ and its animation parameters $\Theta_k$:
\begin{align}
X = \big( (\bm{x}_1, \Theta_1), (\bm{x}_2, \Theta_2), \dots, (\bm{x}_K, \Theta_K) \big) = \big( (\bm{x}_k, \Theta_k) \big)_{k=1}^{K}.
\end{align}
Here, $k$ is the sprite index, $\bm{x}_k \in [0, 1]^{H \times W \times 4}$ is a RGBA texture with of size $H \times W$.
In a more general video decomposition, the texture image is dynamic; \ie, $\bm{x}_k \in [0, 1]^{H \times W \times 4 \times T}$.
However, our main applications of animated graphics often do not include a dynamic texture, so we drop the temporal dynamics in the formulation in this work.
We define the animation parameters $\Theta_k$ by a temporal sequence of tuples of affine warping parameters and a sprite-level opacity parameter: $\Theta_k = (\Theta_k^t)_{t=1}^T = (\bm{a}_k^t, o_k^t )_{t=1}^T$,
where $\bm{a}_k^t \in \mathbb{R}^6$ is affine matrix parameters and $o_k^t \in [0, 1]$ is a sprite-level opacity parameter.

We can render the graphic $X$ into a raster video $Y = ( \bm{y}^t )_{t=1}^T$.
Dropping the notation $t$ for simplicity, at a given time, an RGB frame $\bm{y} \in [0, 1]^{H \times W \times 3}$ is obtained by rendering sprites from the back ($k=1$) to the front ($k=K$):
\begin{align}
  \label{eq:layered2raster1} \bm{b}_1 &= \mathcal{D}(\bm{x}_1; \Theta_1), \\
  \label{eq:layered2raster2} \bm{b}_k &= {\mathcal{B}}(\mathcal{D}(\bm{x}_k; \Theta_k), \bm{b}_{k-1}), \\
  \label{eq:layered2raster3} \bm{y} &= \bm{b}_K,
\end{align}
where $\bm{b}_k$ is an intermediate backdrop of the rendering and $\mathcal{B}$ is the source-over alpha blending function.
The function $\mathcal{D}$ warps the image by affine transform: $\mathcal{D}(\bm{x}; \Theta) = \mathcal{W}([\bm{x}_{\rm RGB}, \bm{x}_{\rm A} \odot \bm{o}]; \bm{a})$,
where $\mathcal{W}$ is an image warping function by a given affine matrix, $[\cdot, \cdot]$ denotes the channel-wise concatenation, %
$\odot$ denotes the element-wise product, and $\bm{o} \in [0, 1]^{H \times W}$ is a 2D array filled with opacity $o$.

\begin{figure}[t]
  \centering
  \includegraphics[keepaspectratio,width=\linewidth]{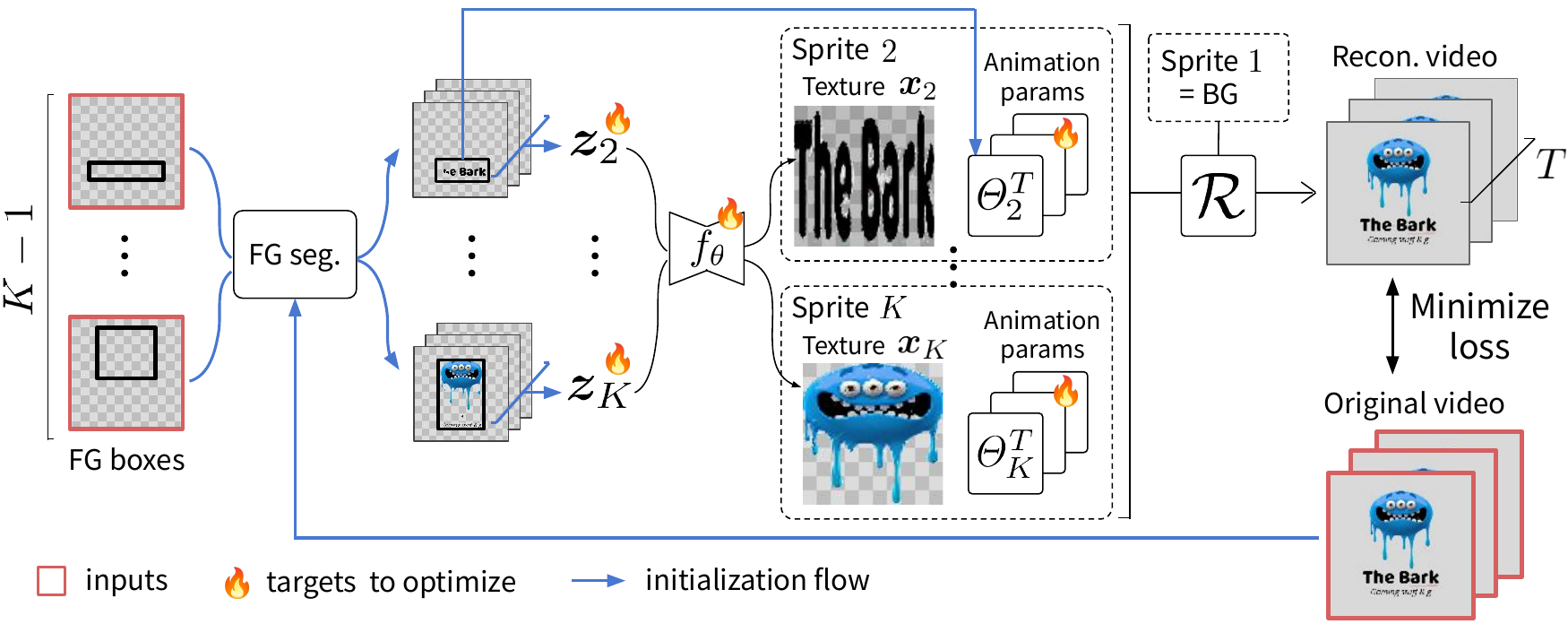}
  \caption{Our decomposition pipeline. Given a raster video and bounding box annotation for a single frame, we first apply a video object segmentation model to initialize texture and animation parameters. Then, we apply a gradient-based optimizer to find the optimal texture codes, animation parameters, and the texture prior parameters.}
  \label{fig:method}
\end{figure}

\subsection{Problem formulation}
\label{sec:problem_formulation}

We define the decomposition problem as finding the optimal parameter $X^\ast$ that can be rendered visually identical to the target raster video $Y$.
Let $\mathcal{L}$ be the function to measure differences between two videos.
The decomposition problem can be expressed by:
\begin{equation}
  \label{eq:optimization}
  \underset{X}{\min} \; \mathcal{L}\big(\mathcal{R}(X), Y\big),
\end{equation}
where $\mathcal{R}(\cdot)$ is the rendering function that applies \cref{eq:layered2raster1,eq:layered2raster2,eq:layered2raster3} frame by frame.
Our experiments use the mean squared error for $\mathcal{L}$.

\section{Approach} \label{sec:method}
While it is possible to apply any optimization approach to the problem in \cref{eq:optimization}, the problem is practically hard to solve due to the complexity of the search space.
For example, naively applying a gradient-based optimizer often results in unwanted artifacts with undesired sprite boundaries.
We employ the following approaches to achieve the good quality/efficiency tradeoff.
1) We introduce an image prior model~\cite{dip} and re-formulate texture optimization as a search for model parameters and codes to prevent undesired artifacts.
2) We slightly simplify the problem setup by assuming the user provides additional auxiliary bounding boxes in a single frame, leading to an efficient initialization method by video object segmentation.
3) We use a robust pre-trained video object segmentation model~\cite{tam} to identify a good initial solution for the optimal sprites.

Given a raster video and auxiliary bounding boxes, we first apply video object segmentation for the foreground sprites and initialize the texture codes and rendering parameters.
Then, we apply a standard gradient-based optimization to find the optimal parameters.
Our simplified representation itself works as a prior and eliminates the need for regularization losses and optimization scheduling as in existing methods~\cite{lna,ds}.

\subsection{Prior-based formulation}
We introduce an image prior model~\cite{dip} to re-formulate our optimization problem (\cref{eq:optimization}).
An image prior model $f_\theta$ represents a mapping of texture code $\bm{z} \in \mathbb{R}^{H \times W \times 4}$
to a texture image $\bm{x}$ with model parameters $\theta$: $\bm{x} = f_\theta(\bm{z})$.
Assuming textures are generated from this prior model, we can transform the texture optimization problem into a search over the code $\bm{z}$ and the model parameters $\theta$.
Let us define $Z = \big( (\bm{z}_k, \Theta_k) \big)_{k=1}^K$.
We can rewrite 
\cref{eq:optimization} in the following:
\begin{align}
  \underset{Z, \theta}{\min} \; \mathcal{L}\big(\mathcal{R}'(Z; \theta), Y\big),
  \label{eq:prior_optimization}
\end{align}
where $\mathcal{R}'$ is the rendering function from $Z$ and the prior model $f_\theta$.

Following the existing study~\cite{ds}, we adopt U-Net~\cite{ronneberger2015u} as a prior model architecture and take the output from the code $\bm{z}$ as the texture.
With multiple sprites, we have an input code $\bm{z}_k$ for each sprite and generate texture $\bm{x}_k$ for each using the shared model parameter $\theta$.
The introduction of the prior model could increase the number of variables in the optimization problem and the time per iteration.
However, we empirically find that the convolutional architecture's inductive bias effectively prevents undesirable artifacts, improves the resulting texture quality, and eventually achieves a good quality/efficiency trade-off.

\subsection{Auxiliary input}
\label{sec:auxiliary}
The transformed formulation of \cref{eq:prior_optimization} is still a hard problem to find a reasonable solution.
In this work, we slightly change the problem setting and assume an additional auxiliary user input that specifies the bounding box of the visible objects in a single video frame.
This auxiliary input tells us 1) the number of sprites to decompose and 2) the rough location of the sprite at time step $\tau$, which allows us to initialize variables effectively.
Our auxiliary input is practically effortless to obtain in interactive video editing, where users are only asked to annotate bounding boxes in a single video frame.

\subsection{Segmentation-based initialization}
\label{subsec:initialization}
The goal of the initialization step is to derive good initial values for $Z$ given the auxiliary user input $((\tau_k, \bm{\beta}_k))_{k=1}^K$, where $\bm{\beta}_k$ is a bounding box for the sprite $k$.
In this work, we initialize the prior parameters $\theta$ by random values~\cite{dip}, and opacity parameters $o_k$ to 1 in all time steps, assuming that all sprites are visible throughout the video frames.

For the initialization of the texture codes $\bm{z}_k$ and affine parameters $\bm{a}_k^t$, we employ an off-the-shelf video object segmentation model~\cite{tam}.
Given the auxiliary user input, we apply a tracking model and obtain bounding boxes and segmentation masks for all time steps $t\neq\tau_k$.
Using the bounding boxes, we initialize the affine parameters $\bm{a}_k^t$ to the box locations with no sheer component.

For initializing textures, we first obtain an initial RGB texture image for each foreground sprite by averaging pixels over time within the bounding box regions.
Similarly, we obtain the alpha channel by averaging segmentation masks over time.
For the background texture ($k=1$), we average the visible pixel values over time for the RGB at each spatial position and set the alpha to 1.
When there are always occluded pixels, we in-paint the region by average pixel values of the visible areas.
Once we obtain the initial texture image, we naively treat them as the initial codes $\bm{z}_k$, which empirically yields good performance. %
After initialization, we apply a standard gradient-based optimization to solve \cref{eq:prior_optimization}.

Our initialization is not perfect due to errors in various sources, such as segmentation, spatial misalignment caused by the texture's deformation, or the transparency effect, but it is still sufficiently effective as the initial solution.
Also, the inference time for the video object segmentation model is negligible compared to the reduction of optimization time thanks to the good initialization.

\subsection{Sprite ordering}
\label{sec:sprite_order}

Our initialization approach has another limitation: our model does not know the order of sprites.
If the rendering order is incorrect, the initial solution may need to swap the order of sprites.
Otherwise, the whole process may fall into a local minimum with a wrong order.
To address this issue, we search for the rendering order that minimizes the reconstruction error for the first $N_{\mathrm{warm}}$ steps of the optimization and then optimize with the rendering order obtained at the $N_{\mathrm{warm}}$-th step.

\section{Crello Animation dataset}
\label{sec:crello_animation_dataset}

We construct a new dataset to study animated graphics.
Inspired by a dataset of static design templates~\cite{yamaguchi2021canvasvae}, we scrape animated templates from the online design service\footnote{\url{https://create.vista.com}} that comes with complete sprite information in each animated graphic.
Our dataset, named the \emph{Crello Animation}\footnote{\url{https://huggingface.co/datasets/cyberagent/crello-animation}} dataset, consists of hundreds of visually appealing animated graphics, mostly designed for social media platforms such as Instagram or TikTok.

In the dataset construction process, we simplify the original templates into the format described in \cref{sec:approach}.
We first export the static image textures for all sprites at the same size.
Then, we compute the animation parameters for each sprite.
The original templates come with sprite animations in one of 17 preset types, such as \texttt{Zoom in/out}, \texttt{Fade in/out}, \texttt{Slide in/out}, and \texttt{Shake}.
If an animation is set on the sprite, we generate per-frame affine matrices and opacities based on their corresponding function.
We apply the identity affine matrix and opacity to all frames if no animation is set.
After converting all the templates, we excluded templates with animated backgrounds as outliers, duplicated templates, and templates with less than two or more than six layers to avoid too complex sprites.
In the end, we obtained 299 samples.
We randomly split them into 154 / 145 samples for validation and test splits. 
More details of the dataset are presented in Appendix.

Compared with natural video datasets such as DAVIS~\cite{davis16} used for video layer decomposition, our Crello Animation has unique characteristics: videos contain various types of texture, including natural images, text, or illustrations, and consist of various numbers of sprites, while the natural scene datasets contain at most two or three objects.
Our dataset preserves complete composition information without artifacts in the background.
This allows us to evaluate the quality of appearance, including occluded areas, while the existing works~\cite{lna,ds} on video layer decomposition have evaluated using Intersection-over-Union of only visible part.
Although several studies~\cite{lamdouar2021segmenting,xie2022segmenting,multiobjectdatasets19,johnson2017clevr} have proposed datasets with composition information like ours, these are synthetic.
Our dataset, sourced from real-world design templates, enables more proper practical evaluations.

\section{Experiments}
\label{sec:evaluation}
We evaluate the performance of our method in decomposing animated graphics and conduct a comparative evaluation with existing video decomposition baselines that output similar layered representations using our Crello Animation.

\subsection{Implementation details}
\label{subsec:implementation_details}

We first tuned the hyperparameters using Crello Animation's validation split and then evaluated their performance on the test split with the tuned hyperparameters.
We used Adam~\cite{kingma2014adam} as the optimizer with a learning rate of $10^{-3}$, set $N_{\mathrm{warm}}$ to $100$ as described in \cref{sec:sprite_order} and set the resolution of $\bm{z}_k$ and textures in our model to $100 \times 100$.
We conducted all experiments on a workstation with a single NVIDIA Tesla T4 accelerator.
We resized the frame size of videos to have a short side of $128$ while keeping the aspect ratio.

For the initialization step, we adopted TAM~\cite{tam} as the video object segmentation model
and used the official implementation\footnote{\url{https://github.com/gaomingqi/Track-Anything}}.
TAM segments a target object by first specifying the target with a user prompt at the keyframe.
We used box prompts and simulated them using ground-truth sprites information in our experiments.
Since in Crello Animation it often happens that most or all of an object's area becomes invisible due to overlapping objects or fading, we calculated the visible area for each sprite, selected the frame with the largest visible area, and generated the rectangle surrounding the visible area in that frame as the prompt.
Our approach to extracting bounding boxes follows the idea that a user annotates a frame where the object has a large visible area.
We obtained the segmentation of the entire video by applying TAM in the forward and backward time directions from the keyframe.
We study the robustness to noise of the user prompt in \cref{sec:box_noise}.

\subsection{Evaluation metrics}
\label{subsec:evaluation_metrics}

We need to measure the quality of the decomposed $\hat{X}$ compared to ground-truth $X$.
An animated graphic $X$ consists of multiple interdependent variables, so the evaluation metrics need to be carefully designed.
We develop multiple evaluation metrics to evaluate the overall quality of the decomposed result.

\noindent{\bf Frame error.}
We measure the reconstruction error between the rendered video frames $\hat{Y}=\mathcal{R}(\hat{X})$ and the original video frames $Y=\mathcal{R}(X)$:
\begin{eqnarray}
  \mathcal{E}_{\rm frame}(\hat{X}, X) = e(\mathcal{R}({\hat{X}}), \mathcal{R}({X})),
\end{eqnarray}
where $e(\cdot,\cdot)$ is a function that measures the error between frames, and we use the mean of pixel-level L1 error or LPIPS~\cite{zhang2018perceptual}.
Since $\hat{Y}$ is reconstructed using all the information of $\hat{X}$, this metric measures the overall quality of the decomposed $\hat{X}$.
However, the frame error alone is insufficient because proximity in the rendered video does not directly reflect the quality of the decomposition result.

\noindent{\bf Sprite error.}
The optimal solution for a sprite $(\hat{\bm{x}}_k, \hat{\Theta}_k)$ is not unique.
For example, two sprites can look identical if the texture is shifted by $1$ pixel, but the affine transformation adjusts the shift by $-1$ pixel.
Thus, we render each sprite and measure the reconstruction error in pixel space.
Also,  in this metric, we aim to measure the quality of each sprite independently, we measure the error after searching for the optimal sprite assignment:
\begin{align}
  \mathcal{E}_{\rm sprite}(\hat{X}, X) = &\underset{\sigma \in S_{K}}{\text{min}} \quad \frac{1}{KT} \sum_{k=1}^{K} \sum_{t=1}^{T}  e(\mathcal{D}(\hat{\bm{x}}_{\sigma(k)}; \hat{\Theta}_{\sigma(k)}^{t}), \mathcal{D}(\bm{x}_k; \Theta_k^{t})),
\end{align}
where $\mathcal{S}_{K}$ represents the set of all possible permutation functions for $K$ elements, with the background always fixed (\ie, $\sigma(1) = 1$).
We define the RGB error by the L1 error weighted by the alpha channel:
\begin{align}
  e_{\rm RGB}(\hat{\bm{y}}, \bm{y}) &= \phi \big(\|\hat{\bm{y}}_{\rm{RGB}} - \bm{y}_{\rm{RGB}}  \|_1 \odot \bm{y}_{\rm A} \big),
\end{align}
where subscripts represent the channels, respectively, and $\phi$ is the average operator over the spatial dimensions.
For the alpha channel, we use the L1 error.

\subsection{Comparison to prior work}
\begin{figure}[t]
  \centering
  \includegraphics[keepaspectratio, width=0.99\linewidth]{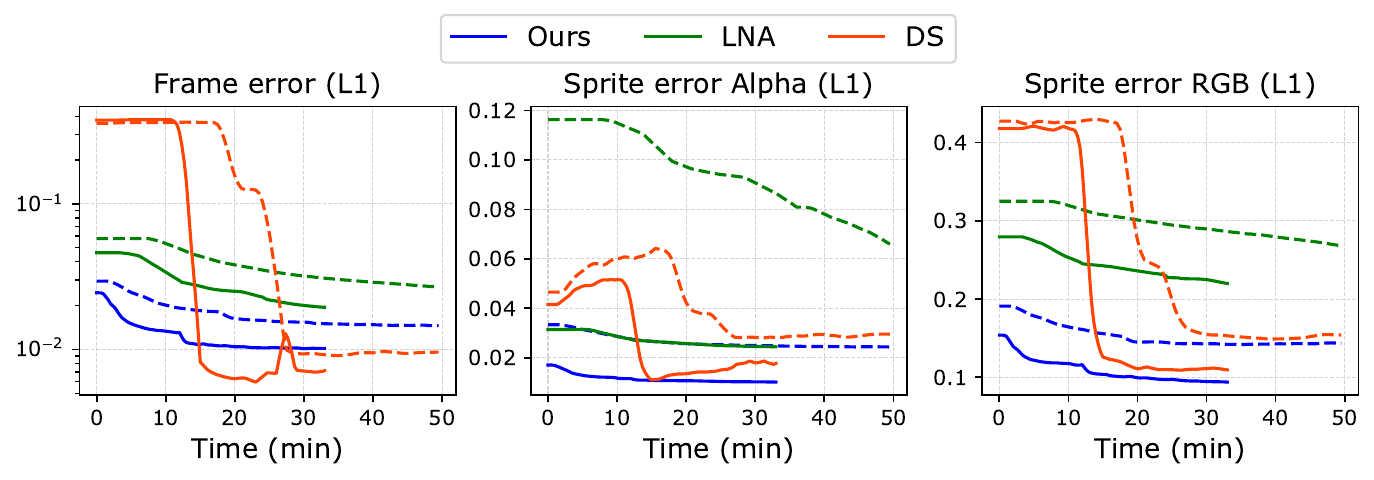}
  \caption{
    Comparison of the trade-off between the quality and optimization time on the test split. 
    The solid lines show the average of the samples with four or fewer layers, and the dashed lines show the average of the samples with five and six layers.
  }
  \label{fig:trade-off}
\end{figure}

\begin{table}[th]
  \caption{
      Quantitative comparison on the test split. All values are averages of the samples. The best and the second best result for each metric are highlighted in {\bf bold} and {\ul underlined}, respectively.
      * indicates the results after optimization convergence.
    }
  \label{tb:comparison_prior_work}
    \centering
    \begin{tabular}{ccccccc}
      \toprule
                                &                            & Time   & \multicolumn{2}{c}{Frame error $\downarrow$} & \multicolumn{2}{c}{Sprite error $\downarrow$} \\
      \multirow{-2}{*}{Method} & \multirow{-2}{*}{\# Iter.} & (min.) & L1                    & LPIPS                & RGB L1                & Alpha L1              \\ \midrule
      LNA                      & 3k                         & 10.4   & 0.0339                & 0.2955               & 0.2654                & 0.0370                \\
      DS                       & 9k                         & 10.6   & 0.2446                & 0.3869               & 0.2965                & 0.0499                \\
      \rowcolor[HTML]{EFEFEF} 
      Ours                     & 11k                        & 10.2   & 0.0163                & 0.0670               & {\ul 0.1179}          & {\ul 0.0193}          \\
      LNA*                     & 11k                        & 40.7   & 0.0163                & 0.1321               & 0.2332                & 0.0336                \\
      DS*                      & 16k                        & 22.8   & \textbf{0.0054}       & \textbf{0.0224}      & 0.1190                & 0.0294                \\
      \rowcolor[HTML]{EFEFEF} 
      Ours*                    & 91k                        & 91.8   & {\ul 0.0101}          & {\ul 0.0411}         & \textbf{0.0984}       & \textbf{0.0179}       \\ \bottomrule
    \end{tabular}
  \end{table}

We compare our method with LNA~\cite{lna} and DS~\cite{ds}, which output similar sprite representations to ours as described in \cref{sec:related_work}.
We used the official implementations of both methods, with slight modifications to fit our setting.
For LNA, we used the TAM's segmentation masks to calculate the mask bootstrapping loss.
For DS, we simplify the transformation as the affine transformation and initialize the parameters in the same way as our method.
We conducted hyperparameter tuning on the validation split and evaluated the performance on the test split for all baselines, including ours.

\cref{fig:trade-off} shows the trade-off curve between optimization time and quality for each method.
Our method shows small errors even at the early stage compared to other methods.
DS receives smaller frame errors than ours as the optimization progresses, but ours is still better in the sprite errors.
We suspect this situation was caused by DS's too-high degree of representation, which can reduce the reconstruction error even if it does not decompose a video well.
Our static texture assumption and limited animation parameters effectively regularize the optimization process and prevent this local minima.

\cref{tb:comparison_prior_work} shows the quantitative comparison on the test split at approximately the same optimization time (10 minutes) and after convergence.
We define the maximum number of iterations for each method as the iteration where the best sprite error is not updated for a quarter of the current iteration on the validation split, and report the errors at the iteration where the loss is minimized as the converged results on the test split (the results on the validation split is in Appendix).
When the optimization time is 10 minutes, our method shows the best results across all metrics, and moreover, it achieves better sprite error than the converged comparative methods.
After convergence, ours achieves the best results in terms of the sprite error.
We emphasize that the frame error is an auxiliary metric that can be low even if the decomposition fails.

\begin{figure}[H]
  \centering
  \includegraphics[width=1.0\linewidth]{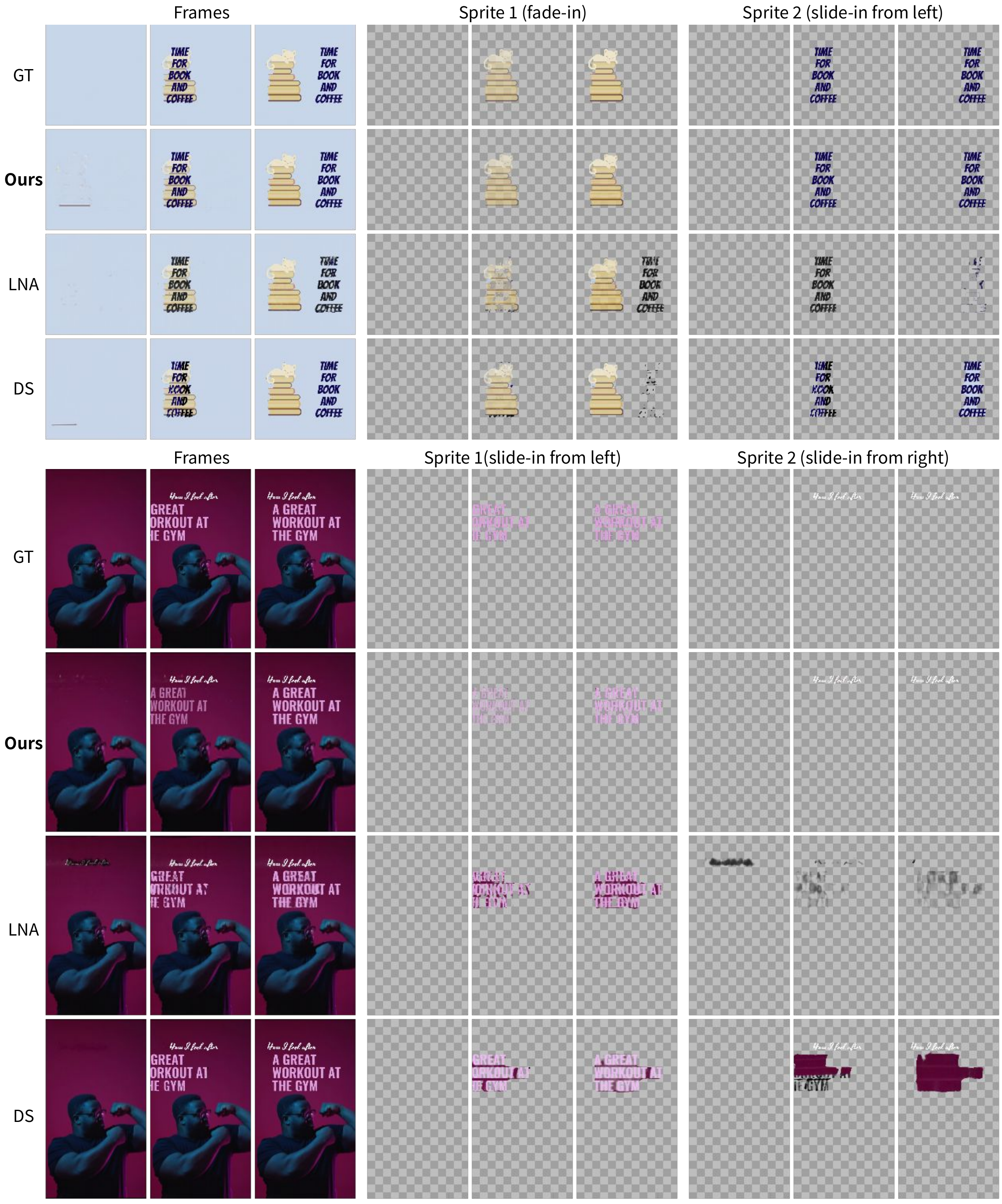}
  \caption{
    Qualitative comparison between LNA~\cite{lna}, DS~\cite{ds}, and our method. 
    We put the description of the animation above each sprite.
  }
  \label{fig:qualitative_comparison}
\end{figure}

Also, we show the qualitative comparison in \cref{fig:qualitative_comparison}.
In LNA, the reconstruction results are generally blurred, and the sprite boundaries are rough.
We suspect LNA has a bias to generate smooth masks since it represents masks with an MLP that tends to output smooth value for the input, \ie coordinate.
DS achieves more precise boundaries than LNA, but DS sometimes fails to group objects.
For example, in the first sample in \cref{fig:qualitative_comparison}, the second sprite is partially included in the first sprite.
Our method does not allow a single sprite to have a complex animation and successfully decomposes this case.
We observe a subtle artifact where the foreground remains in the background.
We suspect this is because the foreground alpha of 1 in the ground truth is not exactly optimized to 1. This causes the occluded area to slightly impact reconstruction, leading to minor RGB inaccuracies that reduce reconstruction loss.
We might be able to rely on post-processing or manual editing since those artifacts often stand out in easily fixable homogeneous regions.
We provide more qualitative results in Appendix.

\begin{table}[t]
  \footnotesize
  \caption{
    Ablation study of our decomposition pipeline. In all settings, the number of total iterations is 20,000. The best and the second best result for each metric are highlighted in {\bf bold} and {\ul underlined}, respectively.
  }
   \centering
   \label{tb:ablation}
  \begin{tabular}{ccccccc}
\toprule
\multicolumn{1}{c}{\multirow{2}{*}{\begin{tabular}[c]{@{}c@{}}Texture\\ prior model\end{tabular}}} & \multicolumn{1}{c}{\multirow{2}{*}{Texture init.}} & \multicolumn{1}{c}{\multirow{2}{*}{Matrix init.}} & \multicolumn{2}{c}{Frame error ↓} & \multicolumn{2}{c}{Sprite error ↓} \\
\multicolumn{1}{c}{}                                                                               & \multicolumn{1}{c}{}                               & \multicolumn{1}{c}{}                              & L1              & LPIPS           & RGB L1          & Alpha L1        \\ \midrule
\checkmark                                                                          & \checkmark                          & \checkmark                         & \textbf{0.0207} & \textbf{0.0793} & \textbf{0.1344} & \textbf{0.0237} \\
                                                                                                   & \checkmark                          & \checkmark                         & 0.0324          & \uline{0.1473}          & \uline{0.1995}          & \uline{0.0283}          \\
\checkmark                                                                          & \checkmark                          &                                                   & 0.0336          & 0.2090          & 0.2651          & 0.0324          \\
\checkmark                                                                          &                                                    &                                                   & \uline{0.0318}          & 0.2258          & 0.3064          & 0.3384          \\ \bottomrule
\end{tabular}
  \end{table}

\subsection{Ablation study}
\label{subsec:ablation_study}
We ablate the effect of each component of our method using the validation split.
We summarize the results in \cref{tb:ablation}.
We can see that using the texture prior improves the decomposition quality in all metrics from the comparison between the first and second rows.
This conforms to the previous report in generative tasks~\cite{dip}.
Comparing rows 3 and 4, texture initialization significantly improves the sprite alpha L1.
This suggests that with no appropriate texture initialization, optimization tends to fall into local minima where the reconstruction error is small, but the sprite is inappropriately decomposed. 
Initializing both affine matrices and textures achieves the best results in all metrics.

\subsection{Robustness to prompt noise} \label{sec:box_noise}
\begin{figure}[t]
  \centering
  \includegraphics[keepaspectratio, width=\linewidth]{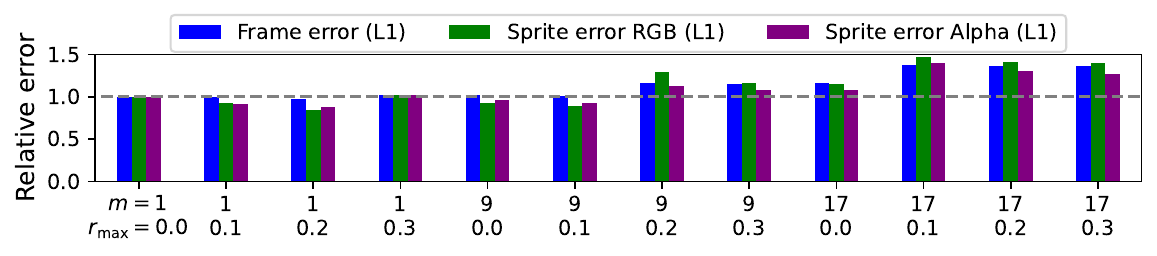}
  \caption{
    Results of evaluating the robustness to prompt's noise. 
    The vertical axis shows the relative error normalized by the one when no noise is added (the leftmost bar).
    The gray dashed line indicates where the relative error becomes 1.
  }
  \label{fig:robustness_to_prompt}
\end{figure}

Although we simulate box prompts by users in our experiments, bounding box annotation usually contains noise in a real-world application.
We verify the robustness of our method to the annotation noise.
We consider two types of noise: noise in the keyframe selection and noise in the box's position and size.
For the former, we select the frame with the $m$-th largest visible area (described in \cref{subsec:implementation_details}) for each foreground sprite, varying the $m$.
For the latter, we add noise to a box directly: $p^{\prime} = p + s \times r$, where $p$ is top, bottom, left, or right coordinate of the box, $s$ is the height if $p$ is the top or bottom and the width otherwise, and $r$ is a random variable sampled from $[-r_{\max}, r_{\max}]$.
We vary the $r_{\rm max}$ and evaluate the decomposition quality.
We show the results in \cref{fig:robustness_to_prompt}.
The results confirm that 
our method does not suffer from critical performance degradation even in the presence of substantial noise ($m = 9$ and $r_{\rm max} = 0.3$),
indicating that users do not need to be nervous about the accuracy of bounding box annotation.

\begin{figure}[t]
  \centering
  \includegraphics[keepaspectratio, width=1.0\linewidth]{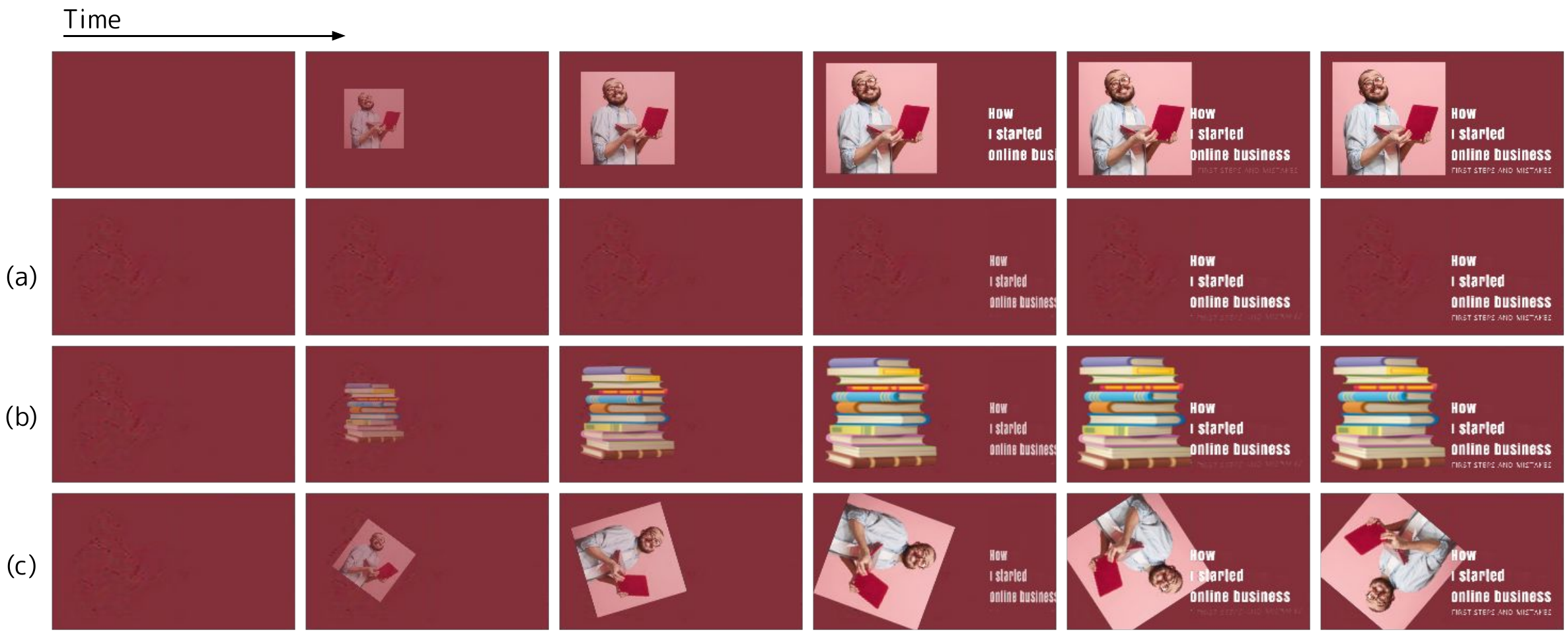}
  \caption{
    Application examples of our decomposition results.
    We decomposed a video in the test split (the first row) using our method and 
    applied three types of editing: (a) sprite removal, (b) texture replacement, and (c) animation (rotation) insertion.
  }
  \label{fig:application}
\end{figure}

\section{Application}
\label{sec:application}
We demonstrate a video editing application using our decomposition approach, shown in \cref{fig:application}.
Here, we first decomposed videos and applied three types of editing: (a) sprite removal, (b) texture replacement, and (c) animation insertion.
In the texture replacement, we replaced the texture of a sprite with a different one while keeping the animation.
In the animation insertion, we added rotation to the original animation while keeping the texture. 
We can observe that the new occlusion and re-appearance caused by the editing are appropriately reflected, and the original animations are correctly transferred to the new textures.
Thanks to our accurate sprite decomposition, these examples do not have major artifacts, which is preferable for video editing.

\section{Conclusion}
\label{sec:conclusion}
We addressed sprite decomposition from animated graphics.
Our optimization-based approach introduces several strategies to make efficient decomposition for the animated graphics, and the evaluation in our newly created Crello Animation shows that our method successfully outperforms existing methods in the trade-off between the quality of the decomposition and the convergence time.

In the future, it would be interesting to relax our static sprite assumption and represent deformation and opacity as a function of time rather than per-frame (\eg, animation representation via keyframes and their interpolation, which is common in video editing software).
This would allow for applications such as increasing the temporal resolution and may improve the performance of the decomposition, functioning as an additional prior.
Also, we are interested in parameterizing videos with more types of animation than motion and opacity, such as blur change and lighting effects, to support creative video workflow.

\bibliographystyle{splncs04}
\bibliography{main}

\newpage
\appendix

\section{Dataset details}
\label{sec:dataset_detail}
\cref{tb:animations} summarizes the description and the number of each animation type in the Crello Animation dataset. 
Each sprite has one of the animation types or no animation.
All animation types can be represented by affine transformation and opacity changes.
In addition to the animation type, each sprite has a delay parameter, which specifies the start time of the animation.
We set the duration of all videos to 5 seconds and adjusted the speed of each animation accordingly, as in the actual rendering engine\footnote{\url{https://create.vista.com}}.
We set the frame rate to 10 for our experiments, but it can be set to any value as the original animations are continuous functions of time.

\begin{figure}[ht]
  \centering
  \begin{subfigure}[b]{0.49\textwidth}
    \centering
    \includegraphics[width=1\linewidth]{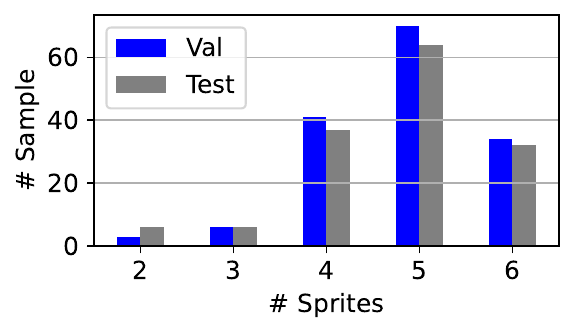}
  \end{subfigure}
  \hfill
  \begin{subfigure}[b]{0.49\textwidth}
    \centering
    \includegraphics[width=1\linewidth]{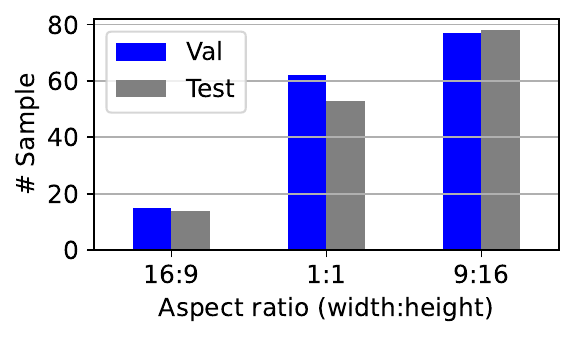}
  \end{subfigure}
  \caption{Statistics in Crello Animation.}
  \label{fig:crello_stats}
\end{figure}

\cref{fig:crello_stats} shows the histogram of the number of sprites in each video and the aspect ratio.
Though the frame resolution can be set to any value by changing the target size of the affine matrices, we set the short edge to 128 pixels for our experiments while keeping the original aspect ratio.

\begin{table}[t]
    \caption{
        Animation types in the Crello Animation dataset. ``None'' indicates that the sprite has no animation.
        }
    \centering
    \begin{tabular}{lC{1.5cm}C{1.5cm}L{7.5cm}}
        \toprule
        Type  & \# Sprites (Val.) & \# Sprites (Test) & Description                                                                                                                                                                                                                                                                                                                            \\ \midrule
        Slide & 244                                                         & 242                                                          & Translation changes continuously. There are three types: \emph{Slide-in}, where translation changes from the start position outside the frame to the base position; \emph{Slide-out}, where translation changes inversely; and \emph{Slide-both}, where both Slide-in and Slide-out occur sequentially. \\ \midrule
        Scale & 165                                                         & 150                                                          & Opacity and scale change continuously. There are three types: \emph{Scale-in}, where opacity and scale change from 0 to 1; \emph{Scale-out}, where opacity and scale change inversely; and \emph{Scale-both}, where both Scale-in and Scale-out occur sequentially.                                     \\ \midrule
        Fade  & 91                                                          & 92                                                           & Opacity changes continuously. There are three types:  \emph{Fade-in}, where opacity changes from 0 to 1, \emph{Fade-out}, where opacity changes inversely, and \emph{Fade-both}, where both Fade-in and Fade-out occur sequentially.                                                                        \\ \midrule
        Zoom  & 58                                                          & 43                                                           & Scale changes continuously. There are three types:  \emph{Zoom-in}, where scale changes from 0 to 1; \emph{Zoom-out}, where scale changes inversely; and \emph{Zoom-both}, where both Zoom-in and Zoom-out occur sequentially.                                                                              \\ \midrule
        Shake & 26                                                          & 12                                                           & Translation oscillates horizontally or vertically.                                                                                                                                                                                                                                                                        \\ \midrule
        Spin  & 6                                                           & 6                                                            & The object rotates around the center axis in the horizontal or vertical direction.                                                                                                                                                                                                          \\ \midrule
        Flash & 4                                                           & 5                                                            & Opacity oscillates between 0 and 1.                                                                                                                                                                                                                                                                                                    \\ \midrule
        None  & 210                                                         & 194                                                          & --                                                                                                                                                                                                                                                                                                                                     \\ \bottomrule
    \end{tabular}
    \label{tb:animations}
\end{table}

\begin{table}[ht]
  \caption{
      Quantitative comparison with prior works on the validation split. 
      All values are averages of the samples. The best and the second best result for each metric are highlighted in {\bf bold} and {\ul underlined}, respectively.
      * indicates the results after optimization convergence.
    }
  \label{tb:comparison_prior_work_val}
    \centering
    \begin{tabular}{ccccccc}
      \toprule
                               &                            & Time   & \multicolumn{2}{c}{Frame error $\downarrow$} & \multicolumn{2}{c}{Sprite error $\downarrow$} \\
      \multirow{-2}{*}{Method} & \multirow{-2}{*}{\# Iter.} & (min.) & L1                    & LPIPS                & RGB L1                & Alpha L1              \\ \midrule
      LNA                      & 3k                         & 10.8   & 0.0308                & 0.2584               & 0.2422                & 0.0271                \\
      DS                       & 9k                         & 10.3   & 0.2771                & 0.4720               & 0.3497                & 0.0369                \\
      \rowcolor[HTML]{EFEFEF} 
      Ours                     & 11k                        & 10.5   & 0.0123                & 0.0510               & 0.1095                & {\ul 0.0116}          \\
      LNA*                     & 11k                        & 40.6   & 0.0145                & 0.1151               & 0.2003                & 0.0214                \\
      DS*                      & 16k                        & 24.4   & \textbf{0.0052}       & \textbf{0.0167}      & {\ul 0.1068}          & 0.0146                \\
      \rowcolor[HTML]{EFEFEF} 
      Ours*                    & 91k                        & 97.9   & {\ul 0.0090}          & {\ul 0.0338}         & \textbf{0.0926}       & \textbf{0.0094}       \\ \bottomrule
      \end{tabular}
  \end{table}

\section{Additional results}
We provide additional quantitative results on the validation split in \cref{tb:comparison_prior_work_val}.
As in the results on the test split in the main paper, our method achieves sprite errors comparable to the converged other methods even in 10 minutes, and achieves even lower sprite errors after convergence.

\begin{figure}[p]
  \centering
  \includegraphics[width=1.0\linewidth]{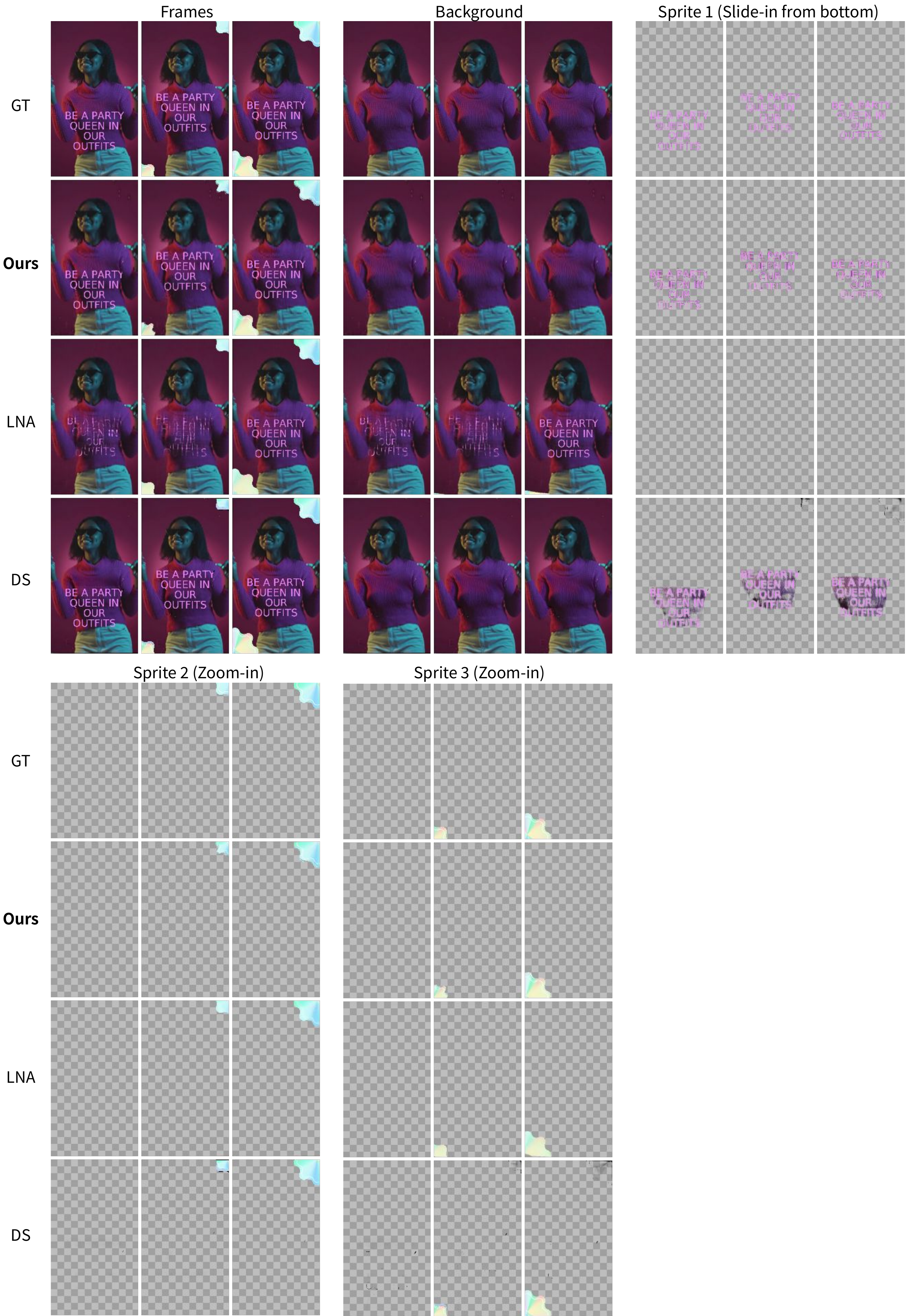}
  \caption{
  Qualitative comparison between Layered Neural Atlases (LNA)~\cite{lna}, Deformable sprites (DS)~\cite{ds}, and our method.
  We put the description of the animation above each sprite.
  Best viewed with zoom and color.
  }
  \label{fig:qualitative_comparison_sup_1}
\end{figure}

\begin{figure}[p]
  \centering
  \includegraphics[width=1.0\linewidth]{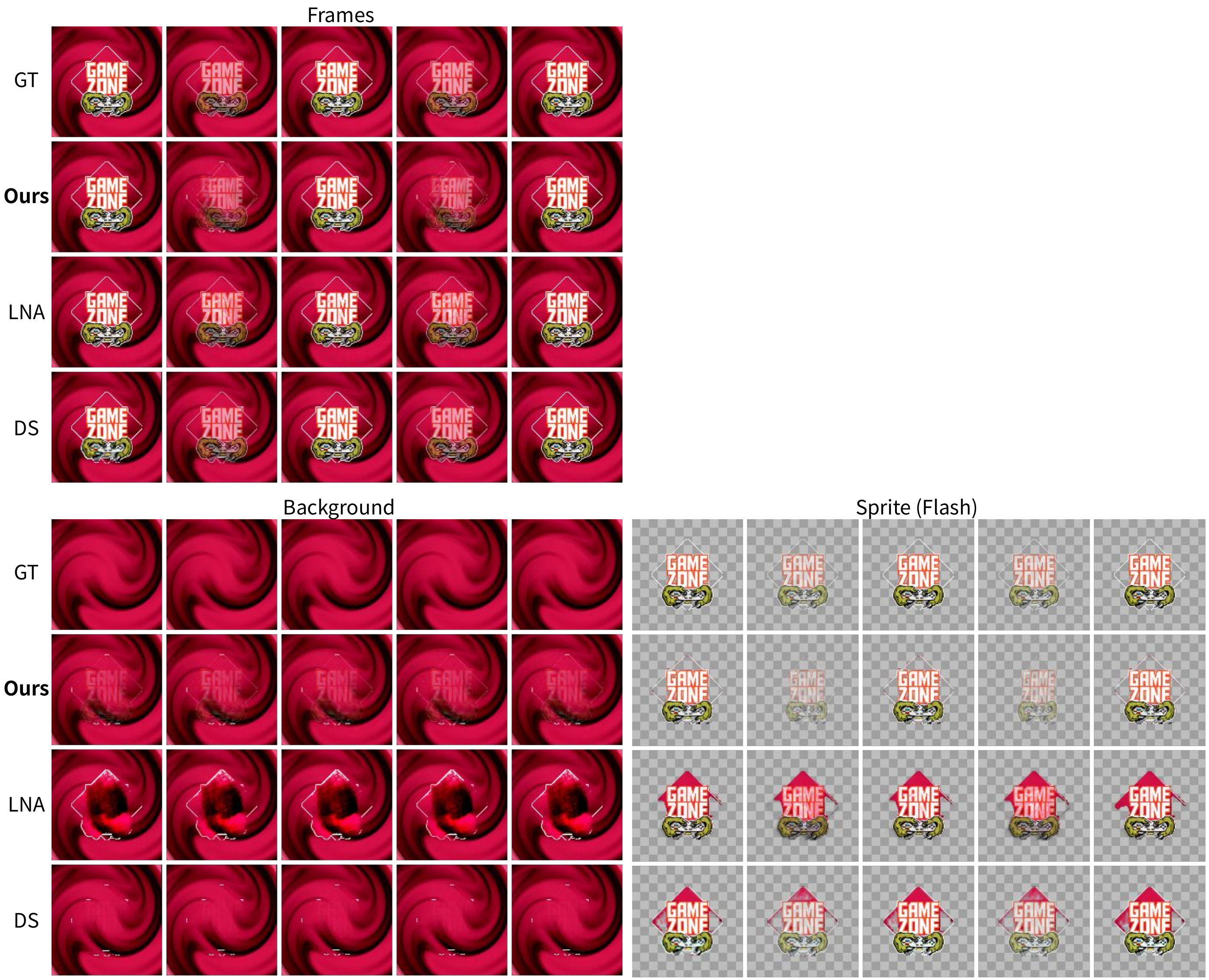}
  \caption{
    Qualitative comparison between Layered Neural Atlases (LNA)~\cite{lna}, Deformable sprites (DS)~\cite{ds}, and our method.
    We put the description of the animation above each sprite.
    Best viewed with zoom and color.
  }
  \label{fig:qualitative_comparison_sup_4}
\end{figure}

\begin{figure}[p]
  \centering
  \includegraphics[width=1.0\linewidth]{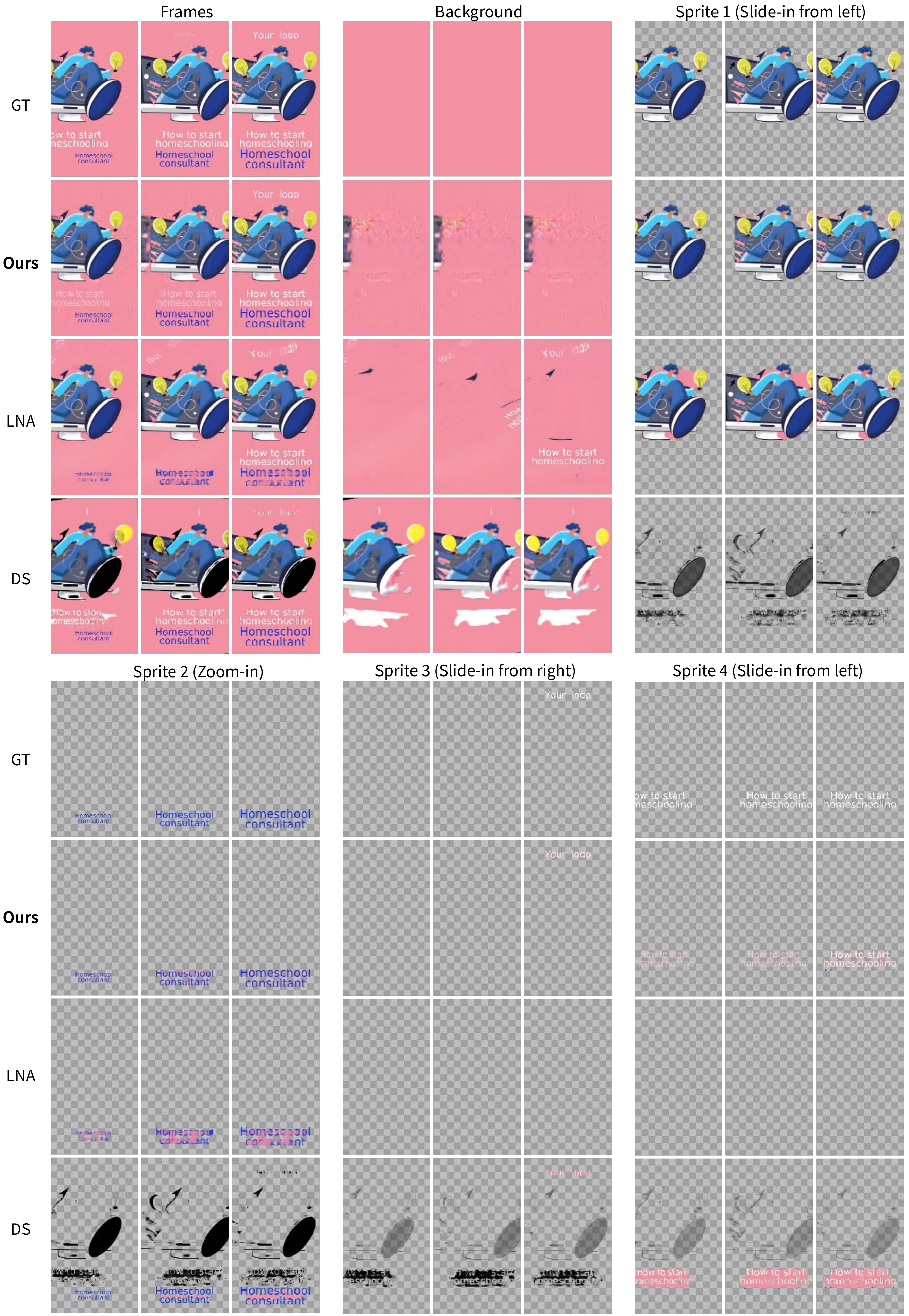}
  \caption{
    Qualitative comparison between Layered Neural Atlases (LNA)~\cite{lna}, Deformable sprites (DS)~\cite{ds}, and our method.
    We put the description of the animation above each sprite.
    Best viewed with zoom and color.
  }
  \label{fig:qualitative_comparison_sup_2}
\end{figure}

\begin{figure}[p]
  \centering
  \includegraphics[width=1.0\linewidth]{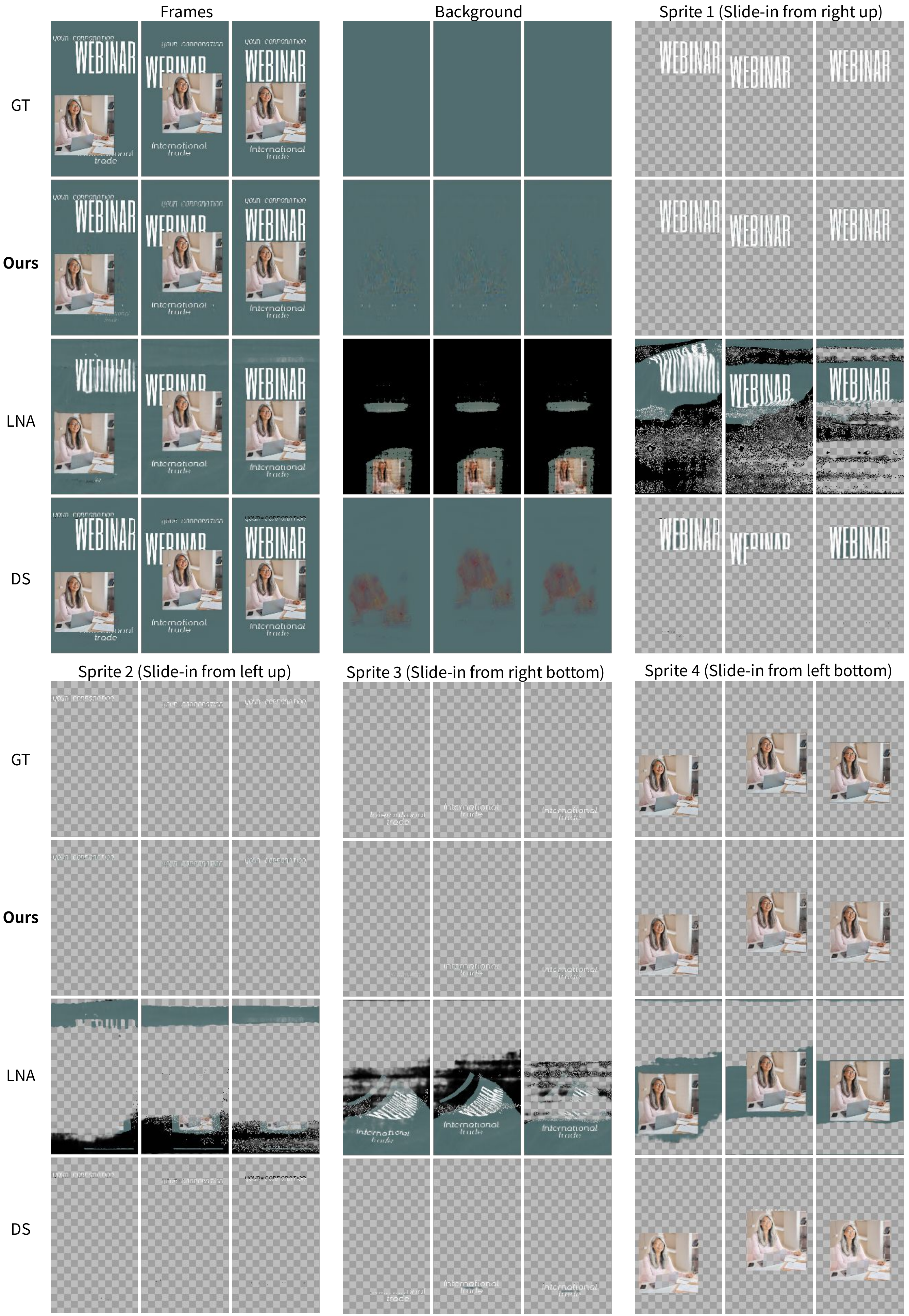}
  \caption{
    Qualitative comparison between Layered Neural Atlases (LNA)~\cite{lna}, Deformable sprites (DS)~\cite{ds}, and our method.
    We put the description of the animation above each sprite.
    Best viewed with zoom and color.
  }
  \label{fig:qualitative_comparison_sup_3}
\end{figure}

\begin{figure}[t]
  \centering
  \includegraphics[width=1.0\linewidth]{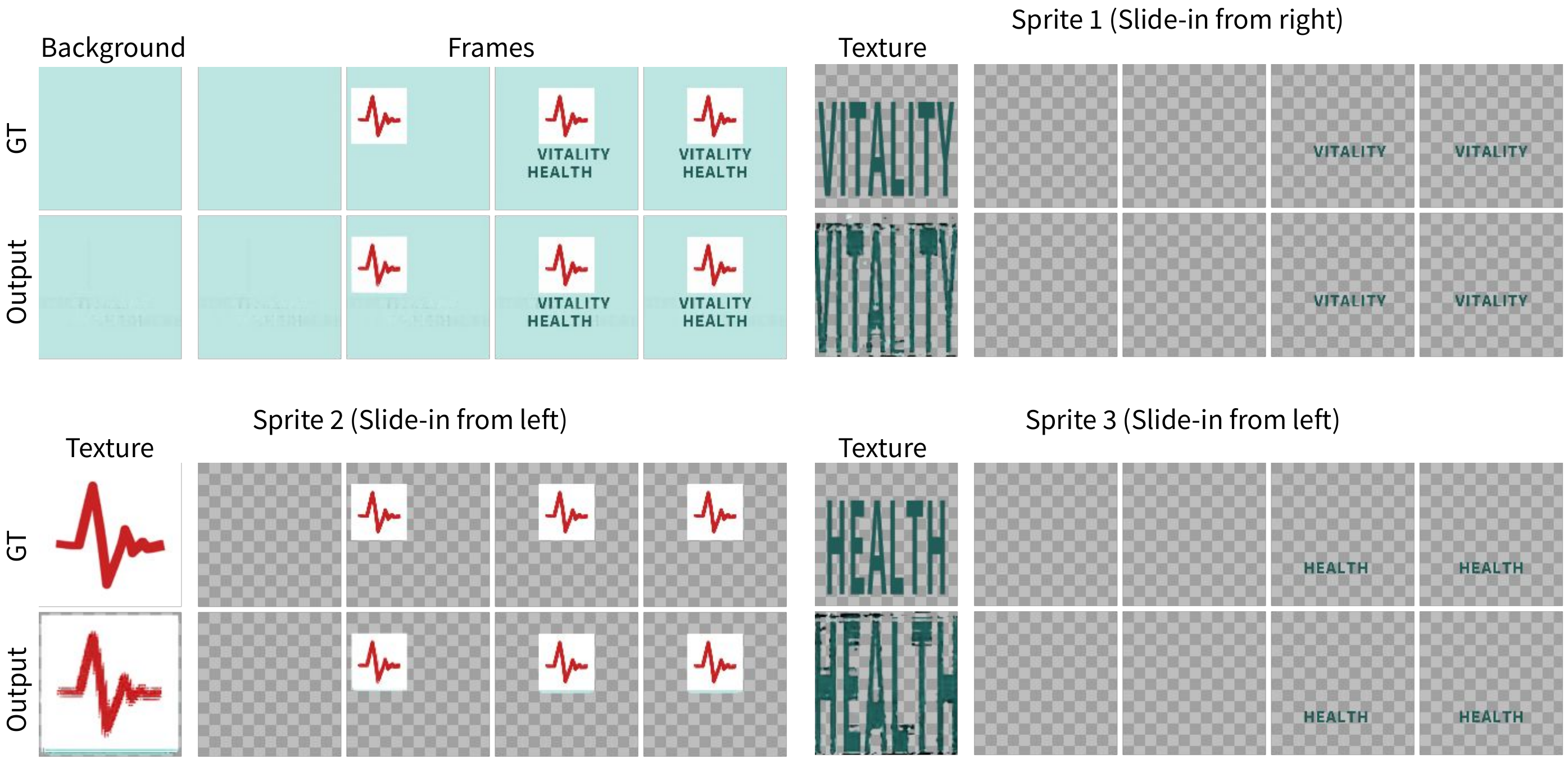}
  \caption{
  Output example of our method. 
  The top-left group shows the background texture and the reconstructed frame and the others show the foreground sprites.
  We put the description of the animation above each sprite.
  Best viewed with zoom and color.
  }
  \label{fig:qualitative_ours_1}
\end{figure}

\begin{figure}[t]
  \centering
  \includegraphics[width=1.0\linewidth]{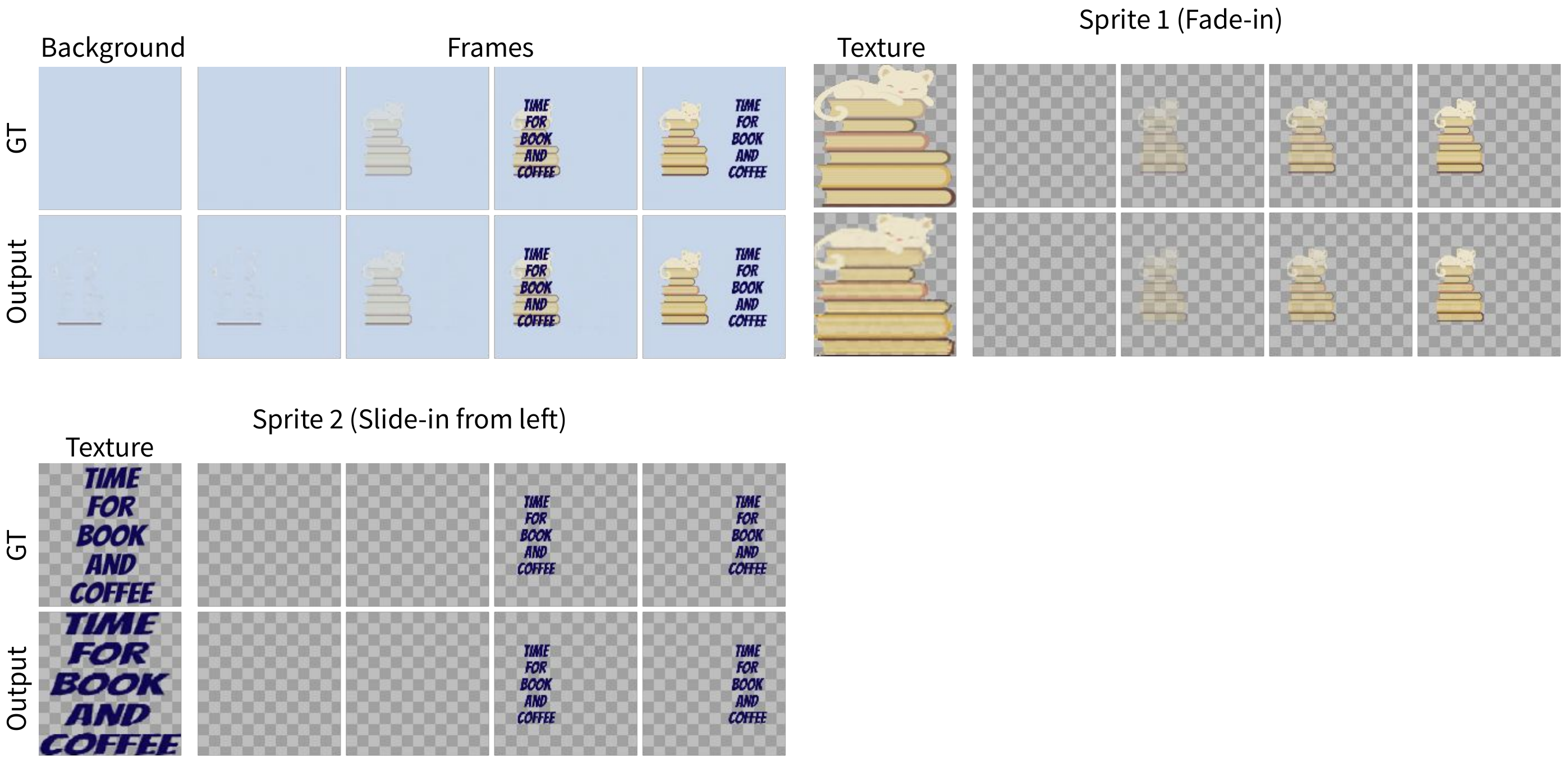}
  \caption{
    Output example of our method. 
    The top-left group shows the background texture and the reconstructed frame, and the others show the foreground sprites.
    We put the description of the animation above each sprite.
    Best viewed with zoom and color.
  }
  \label{fig:qualitative_ours_3}
\end{figure}

We also provide additional qualitative results.
\cref{fig:qualitative_comparison_sup_1,fig:qualitative_comparison_sup_2,fig:qualitative_comparison_sup_3,fig:qualitative_comparison_sup_4} show the comparison between Layered Neural Atlases (LNA)~\cite{lna}, Deformable sprites (DS)~\cite{ds} and our method.
As described in the main paper, our method consistently decomposes sprites with higher quality than LNA and DS, especially for sprites with complex contours such as text.
\cref{fig:qualitative_ours_1,fig:qualitative_ours_3} show more examples of the decomposition results with textures by our method. 
The output textures and ground truth textures may differ in the degrees of freedom of the affine transformation, but the output animations are adjusted accordingly so they are correctly reproduced as sprites.

\begin{figure}[t]
  \centering
  \includegraphics[width=1.0\linewidth]{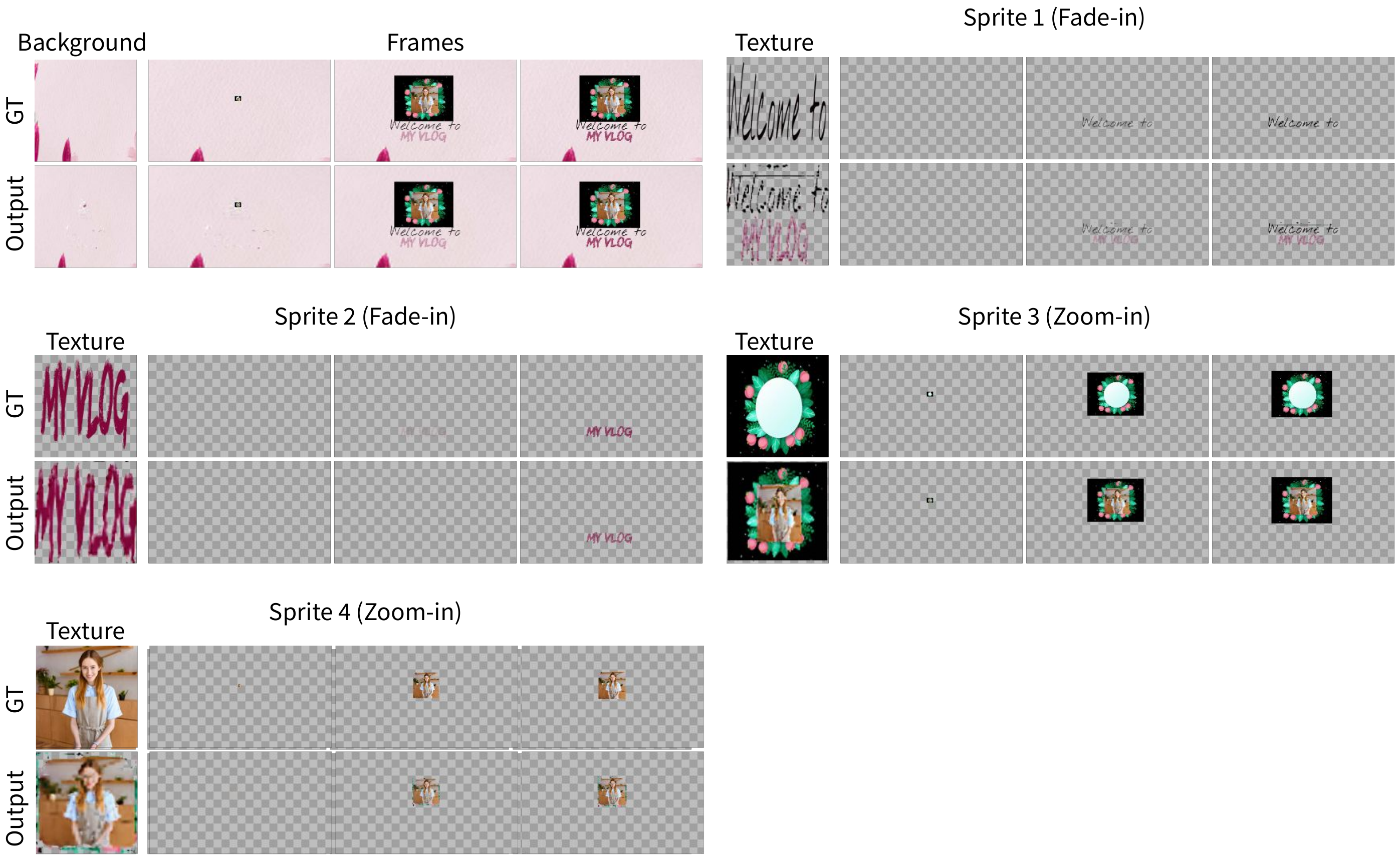}
  \caption{
  A failure case of our method. 
  Sprites with similar animations and close distances are difficult to decompose (as shown in Sprite 1 and Sprite 3).
  The top-left group shows the background texture and the reconstructed frame, and the others show the foreground sprites.
  We put the description of the animation above each sprite.
  Best viewed with zoom and color.
  }
  \label{fig:qualitative_ours_failure_1}
\end{figure}

\begin{figure}[p]
  \centering
  \includegraphics[width=1.0\linewidth]{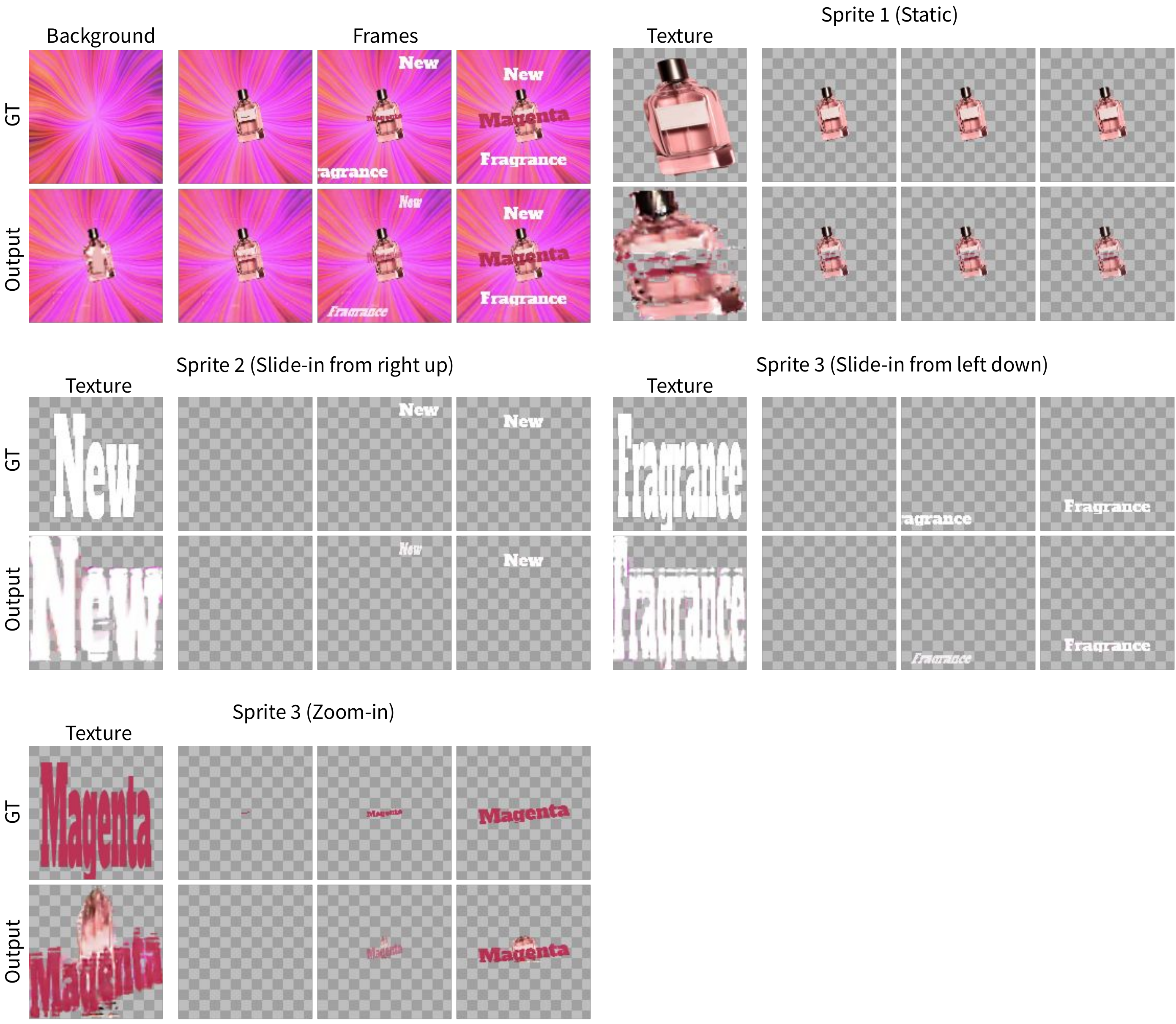}
  \caption{
    A failure case of our method. 
    Sprites with (almost) no animation tend to be absorbed into the background (as shown in Background and Sprite 1).
    The top-left group shows the background texture and the reconstructed frame and the others show the foreground sprites.
    We put the description of the animation above each sprite.
    Best viewed with zoom and color.
  }
  \label{fig:qualitative_ours_failure_2}
\end{figure}

We also show failure cases of our method.
In the first example in \cref{fig:qualitative_ours_failure_1}, the close sprites with similar animations are difficult to decompose.
In the second example in \cref{fig:qualitative_ours_failure_2}, the sprite with (almost) no animation tends to be absorbed into the background.
These failure cases are challenging because they result in small reconstruction errors.
Our initialization should function as a prior to avoid these failures, but further consideration of priors may be necessary.

\section{Baseline details}
We describe the details of the comparison baselines, Layered Neural Atlases (LNA)~\cite{lna} and Deformable Sprites (DS)~\cite{ds}.

\subsection{Layered Neural Atlases}
Based on the official code (updated version)\footnote{\url{https://github.com/thiagoambiel/NeuralAtlases}}, we make a minor modification and tune hyperparameters.
For a fair comparison, we utilize the predicted foreground segmentation masks, which we also used in our method.
Specifically, we add a binary cross-entropy loss to match the predicted alpha with the segmentation mask for each sprite, as the alpha bootstrapping loss in the original paper.
We adopt the original paper's setting except for the weight of the flow alpha loss ($\beta_{f-\alpha}$ in their paper) set to $49$ and the weight of the rigidity loss ($\beta_{r}$ in their paper) set to $1$.
We set the weight of the additional binary cross entropy loss to $10,000$.

\subsection{Deformable Sprites}
We adopted the official implementation\footnote{\url{https://github.com/vye16/deformable-sprites}} to our problem and adjusted several hyperparameters for a fair comparison.
DS uses an image prior model to represent texture images, similar to our method (\$4.1).
We apply the same strategy to initialize the textures as we do (\$4.3).
We also simplify the deformation as the affine transformation and initialize its corresponding parameters in a manner similar to our method. %
To incorporate the given segmentation masks, we employ the binary cross-entropy loss used in LNA to guide the predicted alpha masks.
With a schedule ratio of 1:10 for warm start and main optimization,
we set the weights of the added alpha loss and the dynamic grouping loss ($\mathcal{L}_\mathrm{dynamic}$ in their paper) to 1.0 for the warm start;
we set the weights of the reconstruction loss ($\mathcal{L}_\mathrm{recon}$ in their paper) to 1.0, and the alpha and grouping losses to 0.01 for the main optimization.
We omit other losses, such as optical flow consistency losses, because they do not work effectively in our data domain/problem setting.

\newpage

\end{document}